# Dynamic Impact for Ant Colony Optimization algorithm

Jonas Skackauskas
College of Engineering,
Design and Physical
Sciences
Brunel University London
London, United Kingdom

Tatiana Kalganova
College of Engineering,
Design and Physical
Sciences
Brunel University London
London, United Kingdom
ORCID: 0000-0003-4859-7152

Ian Dear
College of Engineering,
Design and Physical
Sciences
Brunel University London
London, United Kingdom
ORCID: 0000-0003-1289-0402

Mani Janakram
Intel Corporation
Arizona, United States

*Abstract*— This paper proposes an extension method for Ant Colony Optimization (ACO) algorithm called Dynamic Impact. Dynamic Impact is designed to solve challenging optimization problems that has nonlinear relationship between resource consumption and fitness in relation to other part of the optimized solution. This proposed method is tested against complex real-world Microchip Manufacturing Plant Production Floor Optimization (MMPPFO) problem, as well as theoretical benchmark Multi-Dimensional Knapsack problem (MKP). MMPPFO is a non-trivial optimization problem, due the nature of solution fitness value dependence on collection of wafer-lots without prioritization of any individual wafer-lot. Using Dynamic Impact on single objective optimization fitness value is improved by 33.2%. Furthermore, MKP benchmark instances of small complexity have been solved to 100% success rate where high degree of solution sparseness is observed, and large instances have showed average gap improved by 4.26 times. Algorithm implementation demonstrated superior performance across small and large datasets and sparse optimization problems.

*Keywords—Ant Colony Optimization, Dynamic Impact, scheduling, Multi-Dimensional Knapsack Problem, Sparse data*

## I. Introduction

Combinatorial optimization is fundamentally difficult task to be computed. Most of real-world optimization problems are NP-Hard, which mean they are too large to check all combinations in reasonable amount of time to find the optimal result. Instead of brute forcing the optimization, often metaheuristic methods are used to reach "good enough" solution fast. We are using Ant Colony Optimization (ACO) metaheuristic algorithm to solve real world Microchip Manufacturing Plant Production Floor Optimization (MMPPFO) problem and theoretical Multi-Dimensional Knapsack problem (MKP).

The aim of this work is to introduce sub-heuristic search method for Ant Colony Optimization algorithm called Dynamic Impact, and provide insight on how it can be used for any constrained optimization problem. Then use this method to solve real word MMPPFO problem as well as theoretical MKP for further validation and comparison to previous published work.

### A. Ant Colony Optimization

Ant Colony optimization (ACO) is a nature inspired optimization algorithm that uses Ants as search agents navigating a search space. Navigation is mediated by pheromone that ant is naturally draw towards. While ant is navigating it is depositing pheromone on its own path, therefore attracting even more ants. Originally Ant Colony Optimization algorithm has been designed for traveling salesman problem (TSP) described in Dorigo [1] doctoral theses in 1992. ACO algorithm has been successfully implemented to solve number of different problems. Routing problems [2], scheduling and sequencing problems [3] [4] [5] [6] [7], subset problems [8]. Furthermore, ACO has been used to solve large scale optimization problems as demonstrated in these research papers [9] [10] [11].

ACO also has been successfully used for multiple types of scheduling problems. For resource constrained project scheduling problem [3], tardiness problem [4], job shop scheduling problem [5] [6], and other types. Dorigo and Stützle in this research [12] explain implementation methods in detail to solve various scheduling problems using Ant Colony Optimization algorithm.

### B. Microchip manufacturing

Microchip manufacturing is a complex process that utilizes expensive machinery. Tight manufacturing schedules are used in order to run processes at maximum efficiency, minimize machinery down time and always have enough stock of product in demand. Often predicted microchip demand does not meet observed real demand, and microchip production schedule must be altered accordingly to meet newly specified demand.

Microchip manufacturing scheduling problems have been researched from various points of view. Scheduling robotic arms of two-cluster tools in microchip manufacturing facilities [13], transport scheduling in automated material handling systems for wafer manufacturing plants [14], wafer production scheduling as a job shop scheduling problem [15]. In this paper the research focus is on the resource constrained production scheduling.

Microchip manufacturing plant production floor scheduling is a difficult task, as the nature of the problem does not allow to have a set heuristic information on each edge that enables ants to efficiently navigate the search space.

MMPPFO is a scheduling problem. Scheduling problems are proven to be strongly NP-Hard combinatorial problem [16]. To solve scheduling problems variety of different algorithms have been explored. The comparison of Ant Colony Optimization (ACO) versus Genetic Algorithm (GA) and Simulated Annealing (SA) has been conducted buy Huang et. al. [17]. Researchers have found ACO algorithm to be the most effective obtaining feasible solutions for NP-complete scheduling problems. Moreover, semiconductor wafer fabrication scheduling using Ant Colony Optimization was explored in [15] and showed ACO algorithm to be highly effective on a large optimization problem.

## C. Multi-Dimensional Knapsack Problem

In addition to solving real world optimization problem this research further proved the validity of proposed methods by solving a theoretical Multi-Dimensional Knapsack Problem (MKP) as well as compare the results with previously published research papers. The goal of MKP is to maximize the total profit of the items taken into knapsacks, where all items have multi-dimensional weights for each knapsack and each knapsack have a capacity that must not be exceeded [18]. The nature of packing different size items in all knapsacks simultaneously makes the feasible region of the search very sparse [19]. Such sparsity is a great challenge for optimization algorithms where good solutions are obtained by iterative convergence.

## D. Sub-heuristics

Combinatorial search algorithms are designed to explore large search spaces efficiently and converge to a good solution quickly. The efficiency is achieved using metaheuristic methods that allows the search space to be explored more in areas of greater reward. These combinatorial search algorithms usually have multiple hyper parameters, that are often tricky to find such that the best convergence speed is achieved. Hyper-heuristics are methods introduced by Burke et. Al. [20] to generate or choose heuristic that enables combinatorial metaheuristic algorithms converge faster. Hyper-heuristics has been adopted to use multiple low-level heuristic algorithm search results as a search space [21].

For Ant Colony Optimization algorithm such hyper-heuristic usually tunes $\alpha$, $\beta$, and $\rho$ search hyper parameters [22]. However, using similar approaches, it is possible to have more sophisticated search with introduced lower level heuristics within metaheuristic algorithm. The Stochastic Gradient Ascent introduced by Dorigo et. Al. [23] introduced manipulation of the Ant Colony pheromone matrix. Such pheromone correction allows Tuba and Jovanovic [24] avoid algorithm stagnation.

Furthermore, sub-heuristics are the heuristic methods applied within a core of search algorithm that acts upon the state of incomplete partial solution. authors at [25] has utilized such heuristic for Ant Colony Optimization algorithm for probability calculation where a branching can occur while building the solution. This sub-heuristic method allowed them to have transition operation that otherwise could not be accounted from the solutions previously explored.

## E. Data

Our research on MMPPFO has real optimization data for initial testing and capturing real world dynamics provided from industry source. However, to accurately test algorithm performance it is important to have large quantity of data that covers most of optimization scenarios. Also, it is necessary that datasets explore different optimization conditions that could potentially occur in a real optimization system. For effective algorithm evaluation it is beneficial to have methods of creating adjustable difficulty benchmarks [26] by adjusting problem dataset parameters. Furthermore, to fully test a modern optimization algorithm, the benchmark problem has to be challenging enough [27]. Hence, 2 sets of benchmark datasets are chosen to solve MKP problem. Set of small SAC94 datasets and large GK datasets. The datasets are obtained from ResearchGate repository [28].

The rest of this paper is organized as follows. Section II describes the Dynamic Impact method for Ant Colony Optimization algorithm with simplistic usage example. Section III introduces the MMPPFO problem definition, constraints, and objectives. Section IV is dedicated to study of Dynamic Impact method. Study is conducted for two optimization problems MMPPFO and MKP. Also, the exact implementation is of Dynamic Impact is described for each of the problems. Then both problems are solved using Ant Colony Optimization with Dynamic Impact and compared to solutions obtained without Dynamic Impact. Furthermore, for external validation, MKP problem is compared to other recently published research results. Finally, conclusions and future directions are presented in Section V.

## II. ANT COLONY OPTIMIZATION ALGORITHM WITH DYNAMIC IMPACT

### A. Ant Colony Optimization algorithm

Frequently used Min-Max Ant System introduced by Stützle and Hoos [29] is used as a baseline for proposed Dynamic Impact Algorithm extension and experimental work. The algorithm begins with search space initialization in which search space is filter for all nodes to have only feasible edges and each edge pheromone is set to maximum value $\tau_{max}$, also each edge $j$ of the node $i$ gets a precalculated heuristic value $\eta_{j,i}$. Once the search space is prepared, the iterative search starts. In the iterative search a set of ants that each build a complete solution. Each ant starts building with an empty partial solution $s_p = \emptyset$. Then ant searches for a single edge to add to the partial solution. Each edge is added stochastically to the solution using following probability equation:

$$p_{j,i} = \frac{\tau_{j,i}^\alpha * \eta_{j,i}^\beta}{\sum(\tau_{j,i}^\alpha * \eta_{j,i}^\beta)}, \quad \forall(j,i) \in N(s_p) \quad (1)$$

where $\tau$ is edge's pheromone, $\eta$ is edge's heuristic information, $N(s_p)$ is all feasible edges to allowed add to the partial solution $s_p$, $\alpha$ is relative pheromone importance, and $\beta$ is relative heuristic information importance, $j$ and $i$ are the edges and nodes of the search space respectively. Once ant search is finished solution gets evaluated for solution fitness value, and the best solution is passed to influence the global pheromone. At global pheromone update, the pheromone is evaporated using percentage indicated by $\rho$ parameter as in following equation:

$$\tau_{j,i} \leftarrow \tau_{j,i} * (1-\rho), \quad \forall(j,i) \quad (2)$$

where $\rho$ is constant parameter of pheromone evaporation rate introduced by Dorigo and Stützle [30]. Solution of best ant is taken to lay down pheromone on edges that it has visited while building the solution as in following equation:

$$\tau_{j,i} \leftarrow \tau_{j,i} + \rho, \quad \forall(j,i) \in s_p \quad (3)$$

where $s_p$ is the solution of the chosen ant to lay down the pheromone.

The basis of serial algorithm implementation is courtesy of M. Veluscek et al. [31]. However, to utilize modern computer multicore architectures efficiently, parallel ant optimization architecture has been implemented. The parallel ant optimization architecture used in this paper courtesy of I. Dzalbs et al. [32]

This Ant Colony Optimization algorithm is well suited for constrained optimization problems [3]. Heuristic information gives ants sense of direction when pheromone trails are not strong and all edges appear similarly strong in the search space. It plays crucial part in optimization convergence speed. For MMPPFO problem the main objective is minimum undersupported request which it does not have a reliable heuristic information to be pre-calculated, therefore alternative method is needed to provide similar functionality for the search.

The objective of this optimization problem aims to minimize the lack of wafers that has not been scheduled as per order request. Also, producing too many wafers for given request is not productive, and it uses the fab manufacturing capacity that could potentially be used to satisfy other order requests. The order request can be satisfied using collection of wafer-lots. In the supporting collection none of wafer-lots are not more important only the total sum of wafers across all of them. This is a challenge for optimization engines to pick wafer-lots for an order into the partial solution without having clear separation of good wafer-lots versus bad ones, which is normally expressed using heuristic information. This research proposes the Dynamic Impact evaluation method as an extension to Ant Colony Optimization algorithm to improve solution quality and convergence speed.

*B. Dynamic Impact evaluation*

The goal of Dynamic Impact is to enable search identify quicker the good collection of edges for the solution. Dynamic Impact evaluation is novel method of calculating each edge's contribution to fitness value in relation to resource as well as evaluating potential consumption of remaining problem resources before including it to the solution. This allows ant to choose edge more accurately that benefit search fitness value of the solution the most and uses least fraction of resources. This method is a third component in an edge's probability calculation along with pheromone and heuristic information. The Dynamic Impact method is also a myopic search component and it provides search accuracy improvement similar to heuristic information.

Edge's probability calculation using Dynamic Impact:

$$p_{j,i} = \frac{\tau_{j,i}^{\alpha} * \eta_{j,i}^{\beta} * DI_{j,i}^{\gamma}(s_p)}{\sum \left( \tau_{j,i}^{\alpha} * \eta_{j,i}^{\beta} * DI_{j,i}^{\gamma}(s_p) \right)}, \quad \forall (j,i) \in N(s_p) \quad (4)$$

where $DI_{j,i}^{\gamma}(s_p)$ is Dynamic Impact component in probability calculation at the partial solution state $s_p$, $\gamma$ (gamma) is a relative importance of Dynamic Impact, $j$ and $i$ are the edges and nodes of the search space respectively.

Proposed Dynamic Impact component to evaluation is unlike heuristic information and pheromone, this component depends on current state of partial solution and is not pre-calculated like heuristic information. It is designed to change every time an edge is added to a solution. Therefore, it cannot be updated after each solution is completed like pheromone. The best formula for Dynamic Impact calculation depends on optimization problem and optimization goals. Fitness function or simplified version of fitness function is used for calculation of Dynamic Impact. In the cases where fitness function is non-linear relationship of combination of edges the Dynamic Impact measures how much each edge impacts the fitness value for partial solution. Also, it measures the consumption of remaining resources defined as problem constraints in relationship to a reward received from using this edge. General formula of Dynamic Impact can be expressed as following:

$$DI_e = \frac{\left(f(s_p + e) - f(s_p)\right) * A}{\left(\frac{\Omega(s_p) - \Omega(s_p + e)}{\Omega(s_p)}\right)} \quad (5)$$

where $DI_e$ is Dynamic Impact for $e$ edge. $A$ is a sign constant of optimization goal: $+1$ for maximization, and $-1$ for minimization objectives. $f(s_p)$ and $f(s_p + e)$ note the fitness values of a partial solution without and with added edge respectively. Similarly, $\Omega(s_p)$ and $\Omega(s_p + e)$ is a notation of remaining constraints of partial solution without and with added edge respectively. In this theoretical Dynamic Impact calculation, the value is difference of fitness value over the proportion resource consumption. In maximization objective this is similar to perceived value in a given state, where highest increase of fitness may not be the most beneficial if it takes disproportionally large piece of remaining constraints. Parts of this Dynamic Impact function may be simplified depending on an optimization problem. For example, in cases where fitness is linear sum of its solution components $f(s_p + e) - f(s_p)$ can be simplified to just individual fitness of an edge: $f(e)$. Also, constraint part can be simplified whether it has non-linear nature or not, as well as its relevance to the solution. Lastly, the Dynamic Impact formula must always be formulated such that it is always more than zero $DI_e > 0$.

In conclusion, Dynamic Impact evaluation similarly to heuristic information is a myopic search component, however it is evaluated as each edge is added to partial solution, therefore making it more versatile in optimization problems where constant heuristic information value cannot be calculated in advance.

*C. Dynamic Impact example*

Let us consider a simplistic example of vehicle routing where objective is to minimize total time spent on a road for each vehicle but the constraint is fuel in a tank. In such example driving on motorway vehicle might reach the destination faster while using more fuel compared to the more direct route in city traffic that is also much slower. For the purposes of this Dynamic Impact example the formula is simplified to maximize inverse time of the route while using the least portion of remaining fuel.

$$DI_{Route} = \frac{RemainingFuel - FuelCons(Route)}{RemainingFuel * Time(Route)} \quad (6)$$

In Table 1 this formula has been used to demonstrate the difference in Dynamic Impact considering only variable of remaining fuel. In this table there are three routes (edges) to be considered: first fuel efficient but slow, second medium fast and medium fuel efficient, and third fast with high fuel consumption. Three scenarios of remaining fuel are considered: low, medium and high amount of remaining fuel. In scenario one route number one has highest Dynamic Impact because in low fuel situation slow but fuel-efficient route is considered to be more attractive. Second scenario with medium amount of fuel, average fast route is the most attractive. And lastly the third scenario where there is a lot of fuel left to use Dynamic Impact strongly suggest a fastest route. This remaining fuel would not normally be considered in regular ACO probability calculation while building solution and pheromone would have to converge over many iterations without ants having a myopic understanding which of the routes are in their best interest considering the partial solution ant have already built. Using Dynamic Impact ACO can build better solutions from first try and let pheromone continue the fine tuning towards optimal solution along with situation awareness provided by Dynamic Impact.

Table 1: Simplistic example of Dynamic Impact. 3 parallel scenarios that have 3 equivalent routes each. Dynamic Impact is calculated for each route in each scenario individually.

| Scenario | Route number | Route distance | Average route speed | Route time | Fuel consumption | Remaining fuel | Dynamic Impact |
|---|---|---|---|---|---|---|---|
| 1 | **1** | **25** | **10** | **2.5** | **15** | 60 | **0.3** |
|   | 2 | 30 | 15 | 2 | 25 |  | 0.291667 |
|   | 3 | 60 | 60 | 1 | 60 |  | 0 |
| 2 | 1 | 25 | 10 | 2.5 | 15 | 80 | 0.325 |
|   | **2** | **30** | **15** | **2** | **25** |  | **0.34375** |
|   | 3 | 60 | 60 | 1 | 60 |  | 0.25 |
| 3 | 1 | 25 | 10 | 2.5 | 15 | 120 | 0.35 |
|   | 2 | 30 | 15 | 2 | 25 |  | 0.395833 |
|   | **3** | **60** | **60** | **1** | **60** |  | **0.5** |

III. MICROCHIP MANUFACTURING PLANT PRODUCTION FLOOR OPTIMIZATION

Microchip manufacturing plants (fabs) operate continuously all year round according to a planned schedule. When predicted demand is not aligned to actual demand or some unforeseen changes occur, new manufacturing schedules are required to accommodate new demand.
Optimization problem starts with initial wafer-lot production schedule and new die request. To solve the problem, wafer-lot schedule has to be altered to support all demand. Schedule can be altered by changing individual wafer-lot schedule in three major ways: pull-in, push-out, and offload. Pull-in wafer-lot means to produce the wafer-lot earlier. Push-out means to produce the wafer-lot later. Offload means to produce the wafer-lot in another fab. All wafer-lot schedule alterations must comply with existing constraints, therefore making problem combinatorial NP-hard. Wafer production is a complex process in a microchip manufacturing plant. Each fab can produce limited quantity of wafers in selected time window. For the problem solved in this paper, the time window is one Week. With known or predicted future die demand it is possible to create wafer-lot production schedule that maximizes the efficiency of fabs and supports all the requested demand. Moreover, it is desired to support this new demand while having the lowest number of changes to the schedule possible.

A. *Problem definition*

Following are the definitions of MMPPFO are used for this research.
*Wafer-lot* is a non-divisible collection of silicon wafers of a single product to be manufactured all at once and can support only one request. Noted as $WL_i$ where $WL$ stands for wafer-lot and $i$ is the index of the wafer-lot. Each wafer-lot has original schedule slot that can be altered in the problem optimization. For example, wafer-lot $WL_{100}$ can have its commit week changed from $W = 5$ to $W = 3$ which is a pull-in operation as well as at the same time it can be offloaded from $F = F30$ fab to $F = F20$.
*Order* is a silicon wafer product demand to be manufactured in a fab at a specified week. Noted as $O_j$. Where $O$ is stands for order and $j$ is the index of the order. Demand may not be fully satisfied – *undersupported*, or it may have too many wafers scheduled – *oversupported*. For example, order number 5 requests for 55 wafers, $O_5 = 55$. This demand can be supported using multiple wafer-lots.
*Equipped capacity* is a number of wafers of specified product group that a fab is capable to produce at a given week. Noted as $C_{P,F,W}$. Where C stands for capacity, $P$ - product group, $F$ – fab, $W$ - commit week at which the capacity is defined. Specified fab capacity must not be violated as it is physical equipment limitation. For example, $C_{P1,F30,W5} = 400$ is the capacity at fab $F30$ in week $W5$ to make product group $P100$ is 400 wafers. The fab may produce more than one product group and each of them have capacity defined individually. Also, fab capacity is defined for each week individually too, as production capacities might differ week to week.
*Supported request* is a sum of wafers of all wafer-lots that is scheduled to support the request of $O_j$ order

$$SR(O_j) = \sum_i Q(WL_i), \qquad WL_i \in s_p \qquad (7)$$

where $SR(O_j)$ stands for supported request of $O_j$ order, $Q(WL_i)$ is wafer quantity of $WL_i$ wafer-lot, and wafer-lot $WL_i$ belongs to solution where it is used for $O_j$ order.
*Undersupported request* is a number of wafers lacking to support given request in full for $O_j$ Order.

$$USR(O_j) = D(O_j) - SR(O_j) \begin{vmatrix} if\ D(O_j) > SR(O_j) \\ otherwise\ 0 \end{vmatrix} \qquad (8)$$

where $USR(O_j)$ stands for undersupported request of $O_j$ order, $D(O_j)$ is demand of the order.
*Oversupported request* is a number of wafers above the requested demand for $O_j$ order.

$$OSR(O_j) = SR(O_j) - D(O_j) \begin{vmatrix} if\ D(O_j) < SR(O_j) \\ otherwise\ 0 \end{vmatrix} \qquad (9)$$

where $OSR(O_j)$ stands for undersupported request of $O_j$ order.

*Capacity utilization* is a capacity that has been used for wafer production, calculated from an output schedule of an optimization.

$$U(C_{P,F,W}) = \sum_i Q(WL_i), \quad WL_i \in s_p \quad (10)$$

where $U(C_{P,F,W})$ stands for utilization of specified fab capacity $C_{P,F,W}$, and wafer-lot $WL_i$ belongs to solution where it is using fab capacity $C_{P,F,W}$.

*Capacity waste* is a capacity that has been left unused. Capacity waste cannot be negative.

$$WA(C_{P,F,W}) = C_{P,F,W} - U(C_{P,F,W}) \quad (11)$$

where $WA(C_{P,F,W})$ is waste of specified fab capacity $C_{P,F,W}$.

Problem solution noted as $s_p$ is a schedule of wafer-lots to be manufactured. The schedule indicates what wafer-lots $WL_i$ are manufactured at given commit week $W$, and given fab $F$. Fully assembled solution must comply with all problem constraints.

Problem search space noted as $N$ is a collection of all vertices and all edges of feasible combinatorial permutations.

### B. Constraints

This optimization problem has set of constraints that optimization engine must consider simultaneously when building a solution. Some constraints are combinatorial constraints, meaning that combination of wafer-lots must satisfy a given constraint. Other constraints can be search space constraints, that are applied for individual wafer-lot and those constraints limit the total search space to be explored as a consequence.

*1) Capacity constraint*

Fabs have equipped capacity that is a hard limit on how many wafers of specified product group can be scheduled for a given commit week. Sum of wafer must always be lower or equal to equipped capacity. Limit is in effect as a sum of wafers of wafer-lot collection for a given week and fab, thus it is a combinatorial constraint.

$$C_{P,F,W} > U(C_{P,F,W}) \quad (12)$$

*2) Order support constraint*

All wafers supporting an order must be committed on time or ahead of time. This way all wafer-lot permutations that are too late are not included as edges of search space, therefore constraining search space.

$$W(WL_i) \leq W(O_j), \quad \forall (j,i) \in N \quad (13)$$

where $W(WL_i)$ is commit week of $WL_i$, $W(O_j)$ is commit week of $O_j$ order, for all permutations of $j,i$ that belong to search space $N$.

*3) Pull-in, push-out constraint*

Wafer-lot schedule changes must follow specified pull-in push-out information. Pull-in operations for specific products can only be done in fabs that allow to do such operation. Push-out can be done only for a corresponding pull-in operation if necessary to stay within capacity constraint. This constraint limits search space by not including permutations of wafer-lot that has pull-in or push-out operation not defined in the input. Moreover, each push-out must have a corresponding pull-in operation applied in the solution, making it combinatorial constraint too.

*4) Offload constraint*

Each wafer-lot can be offloaded to fabs that support the product group and product itself. This limits search space by not including wafer-lot permutations of offload to fabs that cannot produce product of a wafer-lot.

### C. Optimization objective

In microchip manufacturing, efficiency can be expressed in several different ways. Each solution produced by optimization engine must be evaluated in terms of selected objective to get solution fitness value. Then solution fitness value is compared to other solutions. Solution with lower fitness value is better solution for a minimization objective.

Primary objective of this optimization problem is to minimize undersupported request which makes sure that all customer orders get silicon chips fulfilled on time. Minimizing undersupported request means that all orders should have wafer request supported fully or have least possible number of wafers undersupported.

$$\min \sum_j USR(O_j) \quad (14)$$

where $USR(O_j)$ stands for UnderSupported Request of $O_j$ order.

For new silicon chip demand, it is possible that requested wafers could not be met with integer number of wafer-lots. In such scenario the request will be either undersupported and have the orders not fully complete, or oversupported and waste the production that potentially could be utilized to support other demand.

## IV. EXPERIMENTAL WORK

### A. ACO for Microchip Manufacturing Plant Production Floor Optimization

*1) Search space preparation*

Ants can only navigate efficiently in the prepared search space where all edges are filtered for feasibility and has pheromone and heuristic information value attached to it. In MMPPFO, a wafer-lot possible allocation for production is an edge of a search space. One wafer-lot can have multiple permutations with different production week, and/or production fab.

*2) Heuristic information*

Ant Colony Optimization uses heuristic information that plays very important role in algorithm convergence [33]. Heuristic information gives ants a myopic benefit, and directs them to explore more promising part of search space and obtain good initial solutions before strong pheromone trails are laid. Heuristic information is calculated during search space preparation and remains constant throughout entire algorithm run. However, for MMPPFO problem main objective, minimum undersupported request, does not have an obvious heuristic information that could describe each one of the edges attractiveness separately, since it considers total number of wafers over the collection of multiple wafer-lots taken in the solution. This makes the objective similar to a collection of small subset problems, where individual wafer-lots do not carry any significance over others but only the collection of wafer-lots.

*3) Experimental dataset*

For algorithm validity and performance testing synthetic dataset is needed, covering various corner cases that could potentially occur in real optimization scenario. To generate synthetic datasets an industry provided dataset will be used as initial basis. Aim of dataset generation is to obtain multiple datasets that have dynamics similar to provided dataset, but with added extra desired characteristics and/or features that are not present in dataset provided by industry.

Dataset generation consists of three major parts: wafer-lot generation, order generation, and fab capacity generation.

*Table 2: Dataset parameters for heuristically generated dataset with combinatorial complexity (Wafer-lots x periods) and tightness (total wafer demand / total capacity). Published at figshare: [34]*

| Parameters | Heuristically generated dataset |
|---|---|
| Wafer-lots | 300 (6,312 wafers) |
| Periods (weeks) | 7 |
| Orders | 24 |
| Wafer quantity range | 1-25 |
| Total capacity | 6,000 |
| Total wafer demand | 5,000 |

The generated dataset used in this research in Table 2 is useful for algorithm testing due to increased optimization difficulty by reducing the number of parallel optimal solutions. Accurately measuring number of parallel optimal solutions that exist in search space is an NP-hard question. However, in the context of this optimization problem good difficulty estimation is a ratio of total capacity and wafers over the total wafer demand, which in this dataset is reasonably low.

*4) Dynamic Impact for MMPPFO optimization*

The goal of Dynamic Impact for MMPPFO problem is to enable search quicker and to identify the good collection of wafer-lots to support the order.

Following is description of Dynamic Impact used in optimization for a min undersupported objective:

$$DI_{j,i} = max\{RD(O_j) - |RD(O_j) - Q(WL_i)|, 0.1\} \quad (15)$$
$$RD(O_j) = D(O_j) - SR(O_j) \quad (16)$$

where $DI_{j,i}$ is Dynamic Impact for $O_j$ order and $WL_i$ wafer-lot, $RD(O_j)$ is remaining demand for the $O_j$ order, $D(O_j)$ total demand of the order, and $SR(O_j)$ is supported request of the order. $Q(WL_i)$ is wafer quantity of $WL_i$ wafer-lot. $|RD(O_j) - Q(WL_i)|$ part of formula is a modulus of difference of wafer quantity taken away from remaining demand. $RD(O_j) - |RD(O_j) - Q(WL_i)|$ is that modulus taken from remaining demand. The output of this part when remaining demand is higher than wafer quantity is equal to $Q(WL_i)$. However, if wafer quantity $Q(WL_i)$ is higher than remaining demand then equation gives value lower than $Q(WL_i)$. Lastly the $max$ part of the equation ensures that in the worst case scenario where remaining demand is small enough does not return negative value and as a result algorithm does not calculate negative wafer probability.

This Dynamic Impact evaluation formula represents a simplified edge evaluation fitness function which is the distance of how much each wafer-lot added to solution gets to closer to zero remaining demand and not overshooting. Non-linearity of fitness function is used as a basis of Dynamic Impact for this optimization problem, however it also indirectly represents capacity constraint of a problem too.

Dynamic Impact based purely on fitness function would be the minimum value of remaining demand and wafer quantity, since this would be enough to find accurate difference in fitness value for addition of the wafer-lot to a solution.

*B. Microchip Manufacturing Plant Production Floor Optimization experiment*

The experiment is designed to test the benefit of using Dynamic Impact for Min-Max Ant System in order to achieve best final result. In this experiment, two probability parameters will be tested, $q0$ and $\gamma$. $\gamma$ is the main variable that defines the importance of Dynamic Impact. Experiment baseline is $\gamma = 0$ (Dynamic Impact has no contribution to search probabilities). Moreover, in this experiment $q0$ – the ant exploration hyper parameter is tested, as optimal value of $q0$ often depends on the rest of hyper parameters. $\gamma$ and $q0$ are tested in wide range of values to determine the best possible combination of $\gamma$ and $q0$, as well as to assert the baseline of experiment with $\gamma = 0$ parameter. In this experiment the range of $\gamma$ is from 0.125 growing exponentially to 16 by factor of 2, and $q0$ is from 0.01 growing to linearly to 0.96 by an increment of 0.05.

The remaining parameters of Min-Max Ant System has been established by preliminary experimentation. Best combination of pheromone parameters are: $\tau_{max} = 1$, $\tau_{min} = 0.001$, $\rho = 0.1$. Configuration of probability parameters: $\alpha = 1$, $\beta = 0$. Solutions are achieved running 3,000 iterations using 2 sequential ants, using 16 parallel ants as per [32] described architectural model.

In Table 3 the undersupported score is displayed for each of $q0$ and gamma configuration combinations. Each data is an average score of 20 independent algorithm runs. Firstly, the asserted baseline of $\gamma = 0$, which means Dynamic Impact has no influence on the search probability calculation. Best configuration of $\gamma$ is $\gamma = 0$, $q0 = 0.46$, the corresponding result at this configuration is 30.55 wafers of undersupported request. For the runs which are using Dynamic Impact, the best results are obtained with configuration $\gamma = 4$, $q0 = 0.06$, and the result is 20.4 average wafers of undersupported request score. The difference in best $q0$ value after introduction of Dynamic Impact indicates that algorithm with Dynamic Impact evaluation performs better with higher ant exploration. Using Dynamic Impact with $\gamma = 4$, consistently outperforms $\gamma = 0$ across wider range of $q0$ values. In the context of real-world optimization problems where algorithm $q0$ value is not tuned perfectly, but only roughly estimated $q0$ value. Using the average of 5 best $q0$ settings, at $\gamma = 4$ is 23.08 wafers of undersupported request. In comparison, for imperfectly tuned baseline the average of 5 best $q0$ settings at $\gamma = 0$ is 34.7 wafers of undersupported request.

Moreover, in Fig 1 more detailed comparison of best configurations among baseline $\gamma = 0$, $q0 = 0.46$ and best configuration using Dynamic Impact evaluation $\gamma = 4$, $q0 = 0.06$. In the Fig 1, main bar represents the average undersupported score of 20 algorithm runs of shown setting and error bars indicate one standard deviation of the scores across the runs.

On this optimization problem, with iterations limited to 3,000, using Dynamic Impact evaluation undersupported request score has been improved on average by 33.2%. Moreover, using Dynamic Impact evaluation the standard deviation is reduced from 20.1 to 12.6. This means lower quality solutions occur significantly less often, therefore making performance more reliable, in fast paced environments or solving large scale optimization problems.

Table 3. Undersupported result map for γ and q0, where γ = 0 is an algorithm run without Dynamic Impact. Each data point represents the average of 20 runs. Results of optimizing the heuristically generated dataset.

|  |  | Gamma, γ | | | | | | | |
|---|---|---|---|---|---|---|---|---|---|
|  |  | 0 (no Dynamic Impact) | 0.125 | 0.25 | 0.5 | 1 | 2 | 4 | 8 | 16 |
| q0 | 0.01 | 53.05 | 53.9 | 44.1 | 41.55 | 34.75 | 29.9 | 26.35 | 29.45 | 72.35 |
|  | 0.06 | 50.65 | 49.75 | 46.7 | 34.65 | 29.2 | 23.45 | **20.4** | 33.1 | 63.2 |
|  | 0.11 | 49.3 | 38.6 | 45.35 | 35.6 | 30.35 | 29.45 | 20.55 | 46.9 | 67.45 |
|  | 0.16 | 38.75 | 31 | 35.15 | 36.75 | 28.3 | 25.45 | 39.5 | 35.7 | 57.55 |
|  | 0.21 | 38.4 | 38.05 | 32.7 | 37.45 | 24.55 | 29.8 | 23.1 | 28.65 | 85.1 |
|  | 0.26 | 38.1 | 33.35 | 34 | 35.7 | 30.8 | 22.85 | 25 | 46.75 | 79.4 |
|  | 0.31 | 33.9 | 35.75 | 33.25 | 34.5 | 28 | 32.1 | 42.5 | 60.45 | 85.75 |
|  | 0.36 | 32.55 | 30.95 | 41.25 | 41.6 | 39.95 | 44.65 | 39 | 50.5 | 112.7 |
|  | 0.41 | 43.55 | 54.6 | 49.4 | 60.75 | 48.9 | 56.05 | 47 | 58.3 | 105.3 |
|  | 0.46 | **30.55** | 52.55 | 55.95 | 54.15 | 54.35 | 87.25 | 71.5 | 91.45 | 121.3 |
|  | 0.51 | 44.85 | 59.05 | 47.55 | 82.2 | 72.05 | 87.3 | 80.05 | 88.5 | 132.9 |
|  | 0.56 | 61.05 | 58.45 | 101.4 | 82.1 | 110.4 | 110.6 | 93.15 | 103 | 190.2 |
|  | 0.61 | 66.25 | 86.75 | 113.5 | 122.8 | 120.7 | 119.1 | 122 | 135.9 | 202.1 |
|  | 0.66 | 81.3 | 110.3 | 115.6 | 151.3 | 153.6 | 149.6 | 153.9 | 169.6 | 209 |
|  | 0.71 | 102.8 | 146.1 | 159.2 | 146.1 | 168.8 | 174.6 | 181.5 | 185.7 | 246.2 |
|  | 0.76 | 127.8 | 186.4 | 178.9 | 191.9 | 193.3 | 208.5 | 211.9 | 197.8 | 237.8 |
|  | 0.81 | 177.2 | 202.1 | 217.4 | 205.9 | 209.5 | 242.7 | 230.5 | 239.2 | 243.6 |
|  | 0.86 | 215.6 | 239.3 | 264.9 | 239.6 | 292.1 | 284.9 | 266.2 | 288.8 | 283.8 |
|  | 0.91 | 302.5 | 285.7 | 313.2 | 332.6 | 370.2 | 340.3 | 373.3 | 423.4 | 413.2 |
|  | 0.96 | 396.3 | 390.6 | 408.2 | 436.1 | 443.8 | 488.2 | 444.3 | 549.3 | 535.4 |

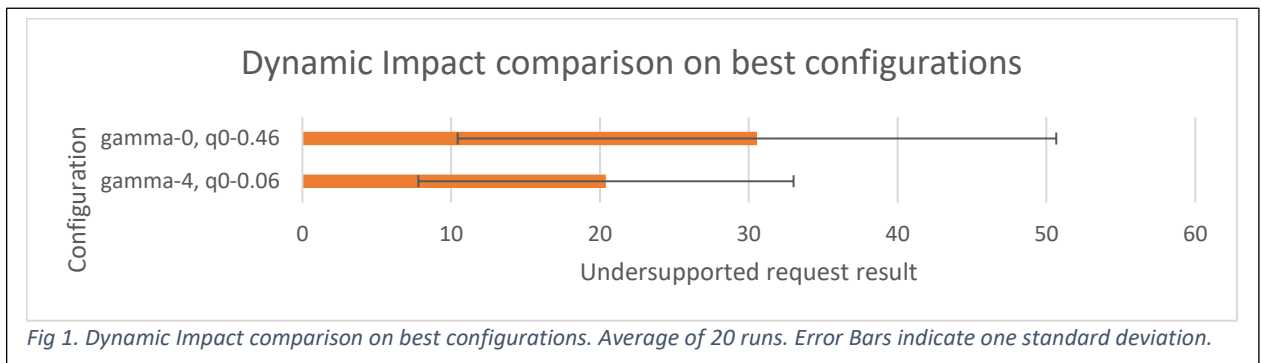

Fig 1. Dynamic Impact comparison on best configurations. Average of 20 runs. Error Bars indicate one standard deviation.

Dynamic Impact evaluation comes with slight computational performance cost. This is due to the fact that Dynamic Impact had to be calculated at each wafer-lot probability calculation. Algorithm at best configuration without Dynamic Impact runs on average 86.8 seconds. With Dynamic Impact using best configuration algorithm took on average 96.9 seconds. This makes Dynamic Impact evaluation add 11.6% of computational overhead. This was possible due to the simplified wafer-lot impact on solution fitness value which made the evaluation not computationally expensive operation.

In conclusion, the method of Dynamic Impact evaluation has proven to be extremely beneficial for objective where the aim is to have combination of elements adding up to the specific requested size or number. This experiment is a meaningful discovery for an algorithm of Ant Colony Optimization that can enable this algorithm to solve broader set of problems with high computational efficiency.

*C. ACO for Multidimensional Knapsack Problem (MKP)*

In addition to solving MMPPFO we have implemented optimization engine to solve MKP too. The purposes of solving MKP is to test Dynamic Impact evaluation method on another benchmark optimization problem.

*1) Search space preparation*

Search space of MKP is simple, MKP does not have discrete orders or nodes and it is a binary problem. The search space is expressed in a single dimension of a binary option, to take item in the knapsack or not. Pheromone $\tau_i$, heuristic information $\eta_i$, and Dynamic Impact $DI_i$ are in this case single dimensional too. The probability calculation for this problem is done all at once for all items before adding any item into partial solution.

*2) Heuritic information*

Similarly, to MMPPFO, MKP maximum profit objective depends on total profit of the collection of all items taken in the knapsack. None of the items are more important in the knapsack over the other, only combination of the items that all simultaneously fit in all knapsack dimensions must have highest profit possible. Therefore, there is no reliable heuristic information available for MKP optimization.

*3) Experimental dataset*

MKP optimization has been chosen in part due to the large availability of benchmark datasets as well as available research publishing the results that state-of-the-art optimization algorithms have achieved. The datasets are obtained from ResearchGate repository [28]. In this paper small SAC-94 datasets and larger GK datasets will be solved. For small SAC-94 datasets, the focus is on achieving optimal values with highest success rate as possible, and on larger GK goal is to get highest profit on average.

*4) Dynamic Impact for MKP optimization*

Dynamic Impact evaluation equation to solve MKP is different from MMPPFO problem as problem domains are not the same. For this problem the Dynamic Impact formula is following:

$$DI_i = \frac{NP(I_i)}{CI(I_i)} \quad (17)$$

$$CI(I_i) = max_{\forall j}\left\{\frac{W(I_i)}{RC(K_j)}\right\} + \frac{\sum_j\left\{\frac{W(I_i)}{RC(K_j)}\right\}}{j} \quad (18)$$

$$NP(I_i) = \frac{P(I_i)}{max\{\forall P(I)\}} \quad (19)$$

$DI_i$ – is Dynamic Impact for item $I_i$, calculated using normalized item profit over capacity impact of the item. Normalized profit $NP(I_i)$ of the item $I_i$ is a constant parameter precalculated using profit of the item and highest profit of all items. It is important to have normalize profit from 0 to 1 in Dynamic Impact such that probability calculations have constant range of inputs for any item profit magnitude range across various input datasets. $CI(I_i)$ is capacity impact of item $I_i$. This is the most intense compute operation of the Dynamic Impact evaluation. It finds the maximum weight utilization combined with average weight utilization of remaining knapsack capacities. Capacity impact has to be recalculated every time doing the probability calculations as it uses the remaining knapsack capacities $RC(K_j)$ in contrast to total capacity that does not change while building the solution. When using remaining knapsack capacity, the current state of solution is well reflected and therefore can have an impact in probability calculation to pick an item that does consume lower portion of available knapsack space for same profit reward. $W(I_i)$ is weight, and $P(I_i)$ is the profit of the item defined in the input dataset.

*D. MKP experiment*

This MKP experiment is chosen in addition to solving MMPPFO problem to solve a commonly available benchmark problem that has similar multiple item collection characteristic. There are no recent papers published on Ant Colony Algorithm solving MKP benchmark datasets, therefore it is logical to assume that there have not been any successful attempts to achieve results on public benchmark datasets to level that is comparable to other published results. Two sets of benchmarks MKP datasets are considered in this experiment. First set SAC-94 are small datasets and are possible to find the optimal solution of a dataset within reasonable amount of time. For these small datasets algorithm success rate is analyzed, and compared which algorithm on average reaches optimal solution quicker. Second set is large GK benchmark datasets. The combinatorial complexity of these benchmark datasets are high enough such that not all of GK datasets have known optimal value, therefore in Table for comparison most recent best known values will be taken from [35] paper that combines their own reached highest values as well as [36] and authors of the GK datasets [37]. The aim for large GK datasets is to get highest possible profit or in other words to minimize profit gap to best known solution.

*1) SAC94 results*

For SAC94 experiment Min-Max Ant System parameters has been tuned with preliminary experimentation. Best combination of pheromone parameters are: $\tau_{max} = 1$, $\tau_{min} = 0.001$, $\rho = 0.1$. Configuration of probability parameters: $\alpha = 1$, $\beta = 0$, $q0 = 0.01$. Solutions are achieved running 3,000 iterations using 2 sequential ants, using 64 parallel ants as per [32] described architectural model. Experiment measures success rate, best successful iteration, average successful iteration, and average profit of each dataset using Dynamic Impact versus algorithm without Dynamic Impact implemented. Each data point is an average of 100 algorithm runs. In Table 4 SAC94 dataset results presented. Ant Colony Optimization using Dynamic Impact preliminary tests showed that the best convergence is achieved using Gamma (γ) value set to 8. ACO with Dynamic Impact shows 100% success rate in every single dataset while same algorithm without Dynamic Impact manages to do so in 41 out of 54 datasets and remaining datasets average 74.7% success rate. Moreover, optimization with Dynamic Impact on average takes just 12.40 iterations and 0.046 seconds to reach optimal value. Without Dynamic Impact on average takes 128.96 iterations and 0.25 seconds to reach optimal on 41 datasets that managed successfully converge 100% of the time.

Our achieved results of SAC94 are compared to recently published research on state of the art optimization algorithms solving SAC94 datasets in Table 5. A binary PSO with time-varying acceleration coefficients (BPSOTVAC) proposed by Chih et. al. [38]. A Dichotomous binary differential evolution (DBDE) proposed by Peng et. al. [39]. A modified version of the flower pollination algorithm (MFPA) proposed by Abdel-Basset et. al. [40]. A binary particle swarm optimization with genetic operations (HPSOGO) introduced by Mingo López et. al. [41]. A random binary differential search algorithm using Tanh function (TR-BDS) introduced by Liu et. al. [42]. A binary artificial algae algorithm (BAAA) introduced by Zhang et. al. [43]. The main comparison metric of all results is success rate. Proposed ACO with Dynamic Impact shows superiority solving small datasets as none of the reviewed algorithms have such versatility in solving all of the datasets reliably to the optimal value 100% of the time. The closest algorithm MFPA that solve on average 99.42% successfully on the datasets published. However, very important to note that this paper [40] does not have complete SAC94 dataset results therefore versatility of the algorithm is not proven since the success rate is unknown of the remaining datasets. Secondly BAAA has 95.2% on average success rate of 48 datasets. 42 out of 48 datasets has reached 100% success rate.

Table 4: MKP SAC94 datasets. Dynamic Impact result comparison of ACO without Dynamic Impact and ACO with Dynamic Impact. Each dataset is result of 100 runs.

| Dataset | Problem size (N x M) | Optimal | ACO without Dynamic Impact | | | | | ACO with Dynamic Impact | | | | |
|---|---|---|---|---|---|---|---|---|---|---|---|---|
| | | | Success rate | Best successful iteration | Average successful iteration | Average time to success (seconds) | Average profit | Success rate | Best successful iteration | Average successful iteration | Average time to success (seconds) | Average profit |
| hp1 | 28 x 4 | 3418 | 0.97 | 3 | n/a | n/a | 3417.58 | 1 | 0 | 0.75 | 0.00308 | 3418 |
| hp2 | 35 x 4 | 3186 | 0.95 | 7 | n/a | n/a | 3185.1 | 1 | 5 | 36.65 | 0.04048 | 3186 |
| pb1 | 27 x 4 | 3090 | 1 | 4 | 334.51 | 0.25203 | 3090 | 1 | 0 | 0.59 | 0.00303 | 3090 |
| pb2 | 34 x 4 | 3186 | 0.97 | 10 | n/a | n/a | 3185.46 | 1 | 0 | 33.87 | 0.03768 | 3186 |
| pb4 | 29 x 2 | 95168 | 1 | 6 | 17.97 | 0.01701 | 95168 | 1 | 0 | 0.71 | 0.00285 | 95168 |
| pb5 | 20 x 10 | 2139 | 1 | 0 | 40.53 | 0.02307 | 2139 | 1 | 0 | 26.5 | 0.01661 | 2139 |
| pb6 | 40 x 30 | 776 | 1 | 4 | 18.68 | 0.01815 | 776 | 1 | 0 | 0.14 | 0.00242 | 776 |
| pb7 | 37 x 30 | 1035 | 0.94 | 10 | n/a | n/a | 1034.47 | 1 | 0 | 4.6 | 0.00853 | 1035 |
| pet2 | 10 x 10 | 87061 | 1 | 0 | 0.08 | 0.00169 | 87061 | 1 | 0 | 8.44 | 0.00514 | 87061 |
| pet3 | 15 x 10 | 4015 | 1 | 0 | 4.02 | 0.00453 | 4015 | 1 | 0 | 0 | 0.00179 | 4015 |
| pet4 | 20 x 10 | 6120 | 1 | 0 | 10.81 | 0.00924 | 6120 | 1 | 0 | 0 | 0.00211 | 6120 |
| pet5 | 28 x 10 | 12400 | 1 | 7 | 13.92 | 0.0177 | 12400 | 1 | 0 | 0 | 0.00195 | 12400 |
| pet6 | 39 x 5 | 10618 | 0.44 | 32 | n/a | n/a | 10610.16 | 1 | 0 | 10.61 | 0.01599 | 10618 |
| pet7 | 50 x 5 | 16537 | 1 | 36 | 249.55 | 0.41771 | 16537 | 1 | 12 | 67.62 | 0.12189 | 16537 |
| sento1 | 60 x 30 | 7772 | 1 | 39 | 319.23 | 0.59452 | 7772 | 1 | 0 | 0.11 | 0.00396 | 7772 |
| sento2 | 60 x 30 | 8722 | 0.65 | 53 | n/a | n/a | 8718.54 | 1 | 0 | 1.94 | 0.01163 | 8722 |
| weing1 | 28 x 2 | 141278 | 1 | 13 | 32.6 | 0.03052 | 141278 | 1 | 0 | 0 | 0.00155 | 141278 |
| weing2 | 28 x 2 | 130883 | 1 | 14 | 36.05 | 0.02862 | 130883 | 1 | 0 | 0 | 0.00163 | 130883 |
| weing3 | 28 x 2 | 95677 | 1 | 6 | 29.44 | 0.01889 | 95677 | 1 | 0 | 0 | 0.00154 | 95677 |
| weing4 | 28 x 2 | 119337 | 1 | 7 | 21.87 | 0.01853 | 119337 | 1 | 0 | 0 | 0.00193 | 119337 |
| weing5 | 28 x 2 | 98796 | 1 | 4 | 18.06 | 0.01286 | 98796 | 1 | 0 | 0 | 0.00164 | 98796 |
| weing6 | 28 x 2 | 130623 | 1 | 11 | 43.77 | 0.03164 | 130623 | 1 | 0 | 0 | 0.00165 | 130623 |
| weing7 | 105 x 2 | 1095445 | 0 | n/a | n/a | n/a | 1095136 | 1 | 4 | 456.14 | 2.06904 | 1095445 |
| weing8 | 105 x 2 | 624319 | 0.03 | 1981 | n/a | n/a | 620481.5 | 1 | 0 | 0.7 | 0.006 | 624319 |
| weish01 | 30 x 5 | 4554 | 1 | 12 | 27.83 | 0.02154 | 4554 | 1 | 0 | 0 | 0.00212 | 4554 |
| weish02 | 30 x 5 | 4536 | 0.91 | 7 | n/a | n/a | 4535.55 | 1 | 0 | 0 | 0.0024 | 4536 |
| weish03 | 30 x 5 | 4115 | 1 | 3 | 21.84 | 0.01619 | 4115 | 1 | 0 | 0 | 0.00211 | 4115 |
| weish04 | 30 x 5 | 4561 | 1 | 1 | 12.33 | 0.0094 | 4561 | 1 | 0 | 0 | 0.0022 | 4561 |
| weish05 | 30 x 5 | 4514 | 1 | 2 | 10.61 | 0.00862 | 4514 | 1 | 0 | 0 | 0.002 | 4514 |
| weish06 | 40 x 5 | 5557 | 1 | 19 | 189.83 | 0.18289 | 5557 | 1 | 0 | 0.08 | 0.00251 | 5557 |
| weish07 | 40 x 5 | 5567 | 1 | 14 | 35.38 | 0.03701 | 5567 | 1 | 0 | 0 | 0.00247 | 5567 |
| weish08 | 40 x 5 | 5605 | 1 | 15 | 37.97 | 0.04175 | 5605 | 1 | 0 | 0 | 0.00254 | 5605 |
| weish09 | 40 x 5 | 5246 | 1 | 18 | 31.22 | 0.02959 | 5246 | 1 | 0 | 0 | 0.00248 | 5246 |
| weish10 | 50 x 5 | 6339 | 1 | 28 | 65.49 | 0.08092 | 6339 | 1 | 0 | 12.08 | 0.01763 | 6339 |
| weish11 | 50 x 5 | 5643 | 1 | 18 | 62.45 | 0.06658 | 5643 | 1 | 0 | 0 | 0.00248 | 5643 |
| weish12 | 50 x 5 | 6339 | 1 | 20 | 56.96 | 0.06909 | 6339 | 1 | 0 | 7.5 | 0.01246 | 6339 |
| weish13 | 50 x 5 | 6159 | 1 | 18 | 35.51 | 0.04445 | 6159 | 1 | 0 | 0 | 0.00263 | 6159 |
| weish14 | 60 x 5 | 6954 | 1 | 27 | 44.24 | 0.06997 | 6954 | 1 | 0 | 0 | 0.00267 | 6954 |
| weish15 | 60 x 5 | 7486 | 1 | 35 | 74.64 | 0.11307 | 7486 | 1 | 0 | 0 | 0.00325 | 7486 |
| weish16 | 60 x 5 | 7289 | 1 | 39 | 545.29 | 0.85691 | 7289 | 1 | 0 | 0.01 | 0.00308 | 7289 |
| weish17 | 60 x 5 | 8633 | 1 | 30 | 78.55 | 0.1655 | 8633 | 1 | 0 | 0 | 0.00374 | 8633 |
| weish18 | 70 x 5 | 9580 | 1 | 52 | 265.71 | 0.614 | 9580 | 1 | 0 | 0.52 | 0.00531 | 9580 |
| weish19 | 70 x 5 | 7698 | 0.93 | 40 | n/a | n/a | 7697.09 | 1 | 0 | 0 | 0.00346 | 7698 |
| weish20 | 70 x 5 | 9450 | 1 | 61 | 398.67 | 0.85951 | 9450 | 1 | 0 | 0 | 0.00387 | 9450 |
| weish21 | 70 x 5 | 9074 | 1 | 44 | 246.19 | 0.50368 | 9074 | 1 | 0 | 0.02 | 0.00369 | 9074 |
| weish22 | 80 x 5 | 8947 | 0.56 | 54 | n/a | n/a | 8939.08 | 1 | 0 | 0 | 0.00391 | 8947 |
| weish23 | 80 x 5 | 8344 | 1 | 44 | 109.6 | 0.24405 | 8344 | 1 | 0 | 0.05 | 0.00383 | 8344 |
| weish24 | 80 x 5 | 10220 | 1 | 74 | 476.98 | 1.34094 | 10220 | 1 | 0 | 0 | 0.00444 | 10220 |
| weish25 | 80 x 5 | 9939 | 0.94 | 71 | n/a | n/a | 9938.17 | 1 | 0 | 0 | 0.00403 | 9939 |
| weish26 | 90 x 5 | 9584 | 0.48 | 71 | n/a | n/a | 9567.44 | 1 | 0 | 0 | 0.00449 | 9584 |
| weish27 | 90 x 5 | 9819 | 1 | 62 | 135.06 | 0.38311 | 9819 | 1 | 0 | 0 | 0.00448 | 9819 |
| weish28 | 90 x 5 | 9492 | 1 | 65 | 421.75 | 1.14258 | 9492 | 1 | 0 | 0 | 0.00442 | 9492 |
| weish29 | 90 x 5 | 9410 | 1 | 73 | 386.61 | 1.03829 | 9410 | 1 | 0 | 0 | 0.00436 | 9410 |
| weish30 | 90 x 5 | 11191 | 1 | 64 | 325.52 | 1.1109 | 11191 | 1 | 0 | 0.01 | 0.00503 | 11191 |

Table 5: SAC94 results comparison with recently published research.

| Dataset | Problem size (N x M) | Optimal | ACO without Dynamic Impact | ACO with Dynamic Impact | BPSOTVAC - [38] 2014 | DBDE - [39] 2017 | MFPA - [40] 2018 | HPSOGO - [41] 2018 | TR-BDS - [42] 2016 | BAAA - [43] 2016 |
|---|---|---|---|---|---|---|---|---|---|---|
| hp1 | 28 x 4 | 3418 | 0.97 | 1 | 0.38 | | 1 | | 0.4 | 0.93 |
| hp2 | 35 x 4 | 3186 | 0.95 | 1 | 0.67 | | | | 0.97 | 0.27 |
| pb1 | 27 x 4 | 3090 | 1 | 1 | 0.46 | | 1 | | 0.5 | 1 |
| pb2 | 34 x 4 | 3186 | 0.97 | 1 | 0.73 | | | | 0.97 | 1 |
| pb4 | 29 x 2 | 95168 | 1 | 1 | 0.91 | | | | 1 | 1 |
| pb5 | 20 x 10 | 2139 | 1 | 1 | 0.84 | | 1 | | 0.8 | 1 |
| pb6 | 40 x 30 | 776 | 1 | 1 | 0.5 | | 1 | | 0.57 | 1 |
| pb7 | 37 x 30 | 1035 | 0.94 | 1 | 0.47 | | 1 | | 0.8 | 1 |
| pet2 | 10 x 10 | 87061 | 1 | 1 | | | | 1 | | |
| pet3 | 15 x 10 | 4015 | 1 | 1 | | | | | | |
| pet4 | 20 x 10 | 6120 | 1 | 1 | | | | | | |
| pet5 | 28 x 10 | 12400 | 1 | 1 | | | | | | |
| pet6 | 39 x 5 | 10618 | 0.44 | 1 | | | | | | |
| pet7 | 50 x 5 | 16537 | 1 | 1 | | | | | | |
| sento1 | 60 x 30 | 7772 | 1 | 1 | 0.57 | 0.43 | 1 | 0.16 | 0.8 | 1 |
| sento2 | 60 x 30 | 8722 | 0.65 | 1 | 0.27 | 0 | 1 | 0.25 | 0.73 | 1 |
| weing1 | 28 x 2 | 141278 | 1 | 1 | 1 | 1 | | 0.1 | 1 | 1 |
| weing2 | 28 x 2 | 130883 | 1 | 1 | 1 | 0.97 | | 1 | 1 | 1 |
| weing3 | 28 x 2 | 95677 | 1 | 1 | 0.92 | 0.6 | 1 | 1 | 0 | 1 |
| weing4 | 28 x 2 | 119337 | 1 | 1 | 1 | 1 | 1 | 1 | 1 | 1 |
| weing5 | 28 x 2 | 98796 | 1 | 1 | 1 | 0.3 | | 1 | 0.7 | 1 |
| weing6 | 28 x 2 | 130623 | 1 | 1 | 0.97 | 0.97 | 1 | 1 | 1 | 1 |
| weing7 | 105 x 2 | 1E+06 | 0 | 1 | 0 | 0 | | 1 | 0 | 0.58 |
| weing8 | 105 x 2 | 624319 | 0.03 | 1 | 0.35 | 0 | | 1 | 0.5 | 0.93 |
| weish01 | 30 x 5 | 4554 | 1 | 1 | 1 | 1 | 1 | 1 | 1 | 1 |
| weish02 | 30 x 5 | 4536 | 0.91 | 1 | 0.64 | 1 | 1 | 1 | 1 | 1 |
| weish03 | 30 x 5 | 4115 | 1 | 1 | 0.99 | 1 | 1 | 1 | 1 | 1 |
| weish04 | 30 x 5 | 4561 | 1 | 1 | 1 | 1 | 1 | 1 | 1 | 1 |
| weish05 | 30 x 5 | 4514 | 1 | 1 | 1 | 1 | 1 | 1 | 1 | 1 |
| weish06 | 40 x 5 | 5557 | 1 | 1 | 0.59 | 0.3 | 1 | 1 | 1 | 1 |
| weish07 | 40 x 5 | 5567 | 1 | 1 | 0.96 | 0.33 | 1 | 1 | 0.98 | 1 |
| weish08 | 40 x 5 | 5605 | 1 | 1 | 0.79 | 0.87 | 1 | 1 | 0.98 | 1 |
| weish09 | 40 x 5 | 5246 | 1 | 1 | 1 | 1 | 1 | 1 | 1 | 1 |
| weish10 | 50 x 5 | 6339 | 1 | 1 | 0.91 | 1 | 1 | 1 | 1 | 1 |
| weish11 | 50 x 5 | 5643 | 1 | 1 | 0.88 | 0.63 | 1 | 1 | 0.92 | 1 |
| weish12 | 50 x 5 | 6339 | 1 | 1 | 0.89 | 1 | 0.82 | 1 | 0.96 | 1 |
| weish13 | 50 x 5 | 6159 | 1 | 1 | 1 | 1 | 1 | 0.35 | 0.98 | 1 |
| weish14 | 60 x 5 | 6954 | 1 | 1 | 0.98 | 1 | 1 | 1 | 0.92 | 1 |
| weish15 | 60 x 5 | 7486 | 1 | 1 | 1 | 1 | 1 | 1 | 0.96 | 1 |
| weish16 | 60 x 5 | 7289 | 1 | 1 | 0.54 | 0.87 | 1 | 1 | 1 | 1 |
| weish17 | 60 x 5 | 8633 | 1 | 1 | 1 | 0.67 | | 1 | 1 | 1 |
| weish18 | 70 x 5 | 9580 | 1 | 1 | 0.75 | 1 | | 1 | 0.98 | 1 |
| weish19 | 70 x 5 | 7698 | 0.93 | 1 | 0.65 | 1 | 1 | 1 | 0.96 | 1 |
| weish20 | 70 x 5 | 9450 | 1 | 1 | 0.78 | 1 | 1 | 1 | 0.96 | 1 |
| weish21 | 70 x 5 | 9074 | 1 | 1 | 0.74 | 1 | 1 | 0.1 | 0.96 | 1 |
| weish22 | 80 x 5 | 8947 | 0.56 | 1 | 0.16 | 1 | | 1 | 0.98 | 1 |
| weish23 | 80 x 5 | 8344 | 1 | 1 | 0.85 | 0.23 | | 1 | 0.92 | 0.45 |
| weish24 | 80 x 5 | 10220 | 1 | 1 | 0.7 | 1 | | 1 | 0.68 | 0.54 |
| weish25 | 80 x 5 | 9939 | 0.94 | 1 | 0.49 | 0.97 | | 1 | 0.84 | 1 |
| weish26 | 90 x 5 | 9584 | 0.48 | 1 | 0.36 | 1 | 1 | 1 | 0.94 | 1 |
| weish27 | 90 x 5 | 9819 | 1 | 1 | 0.99 | 0.97 | | 1 | 0.98 | 1 |
| weish28 | 90 x 5 | 9492 | 1 | 1 | 0.87 | 1 | | 1 | 0.94 | 1 |
| weish29 | 90 x 5 | 9410 | 1 | 1 | 0.86 | 1 | | 1 | 0.92 | 1 |
| weish30 | 90 x 5 | 11191 | 1 | 1 | 0.87 | 0.83 | | 1 | 0.32 | 1 |

None of the authors has considered "pet" datasets part of SAC94. "pet" datasets seem to be an edge case, especially problematic for any optimization algorithm with observed highly sparse nature, and despite small theoretical combinatorial complexity, and are difficult to solve. None of the other research has published results solving "pet" datasets, possibly due to difficulty handling high degree of sparseness, especially when it is expected to be easily solved as theoretical combinatorial complexity is low.

*1) GK results*

Algorithm has been tuned slightly differently to solve large GK datasets. Dynamic Impact importance parameter Gamma (γ) value is set to 32, and algorithm is run for 10000 iterations. Experiment measures average profit obtained over 10 algorithm runs, then average profit is turned into average gap using best known profit values. In Fig 2. ACO with Dynamic Impact is compared to the same algorithm without implemented Dynamic Impact running the same probability settings. In absolute terms ACO with Dynamic Impact gets average gap reduction of 0.54%, where highest difference is in gk09 – 0.9% and lowest in gk01 – 0.27%. In relative terms difference in profit gap is on average 4.27 times lower, where highest is gk02 reducing gap 10.4 times and lowest in gk03 reducing gap 2.33 times. Furthermore, in Fig 3. well performing ACO with Dynamic Impact algorithm is stacked up against recently published solutions of GK dataset implementations. Dantas – GPGPU SA [44] is GPU accelerated Simulated Annealing algorithm. Kong – NBHS2 [45] out of several algorithms compared their proposed New Binary Harmony Search type 2 was best performing for GK datasets. Wang – DLHO [46] is their proposed Diverse Human Learning Optimization algorithm that has performed the best among compared solutions. On average ACO with Dynamic Impact has 0.31% or 3.3 times lower gap than Dantas – GPGPU SA, however ACO is outperformed by 0.05% difference in gap on single gk09 instance. Kong – NBHS2 has closer performance and is on average 0.24% or 2.48 times behind ACO, however no instances outperform ACO, and the closest instance is gk07 falling behind by 0.07% or 1.35 times. Lastly, ACO outperforms Wang – DLHO on average by 1.10% or 7.72 times.

In conclusion Dynamic Impact proved to significantly aid the search for small datasets reliably reach optimal value and large datasets significantly lower gap to optimal value.

## V. CONCLUSIONS

This research has studied Ant Colony Optimization algorithm solving MMPPFO. Problems main optimization objective depends on a collection of smaller parts of solution without prioritizing any one over others therefore useful heuristic information that could be predefined does not exist. The research has proposed additional component to the ACO algorithm probability calculation which is called Dynamic Impact. Dynamic Impact similarly to heuristic information is a myopic component of the search. The difference is, Dynamic Impact is calculated each time probability is calculated and it depends on a state of partial solution. In other words, Dynamic Impact is simplified evaluation of each edge impact on fitness function and resource consumption. Computational overhead to use this method is low when micro-optimized for specific problem. For MMPPFO problem, this research has demonstrated that using Dynamic Impact evaluation significantly improve solution quality over the same number of search iterations. Furthermore, ACO with

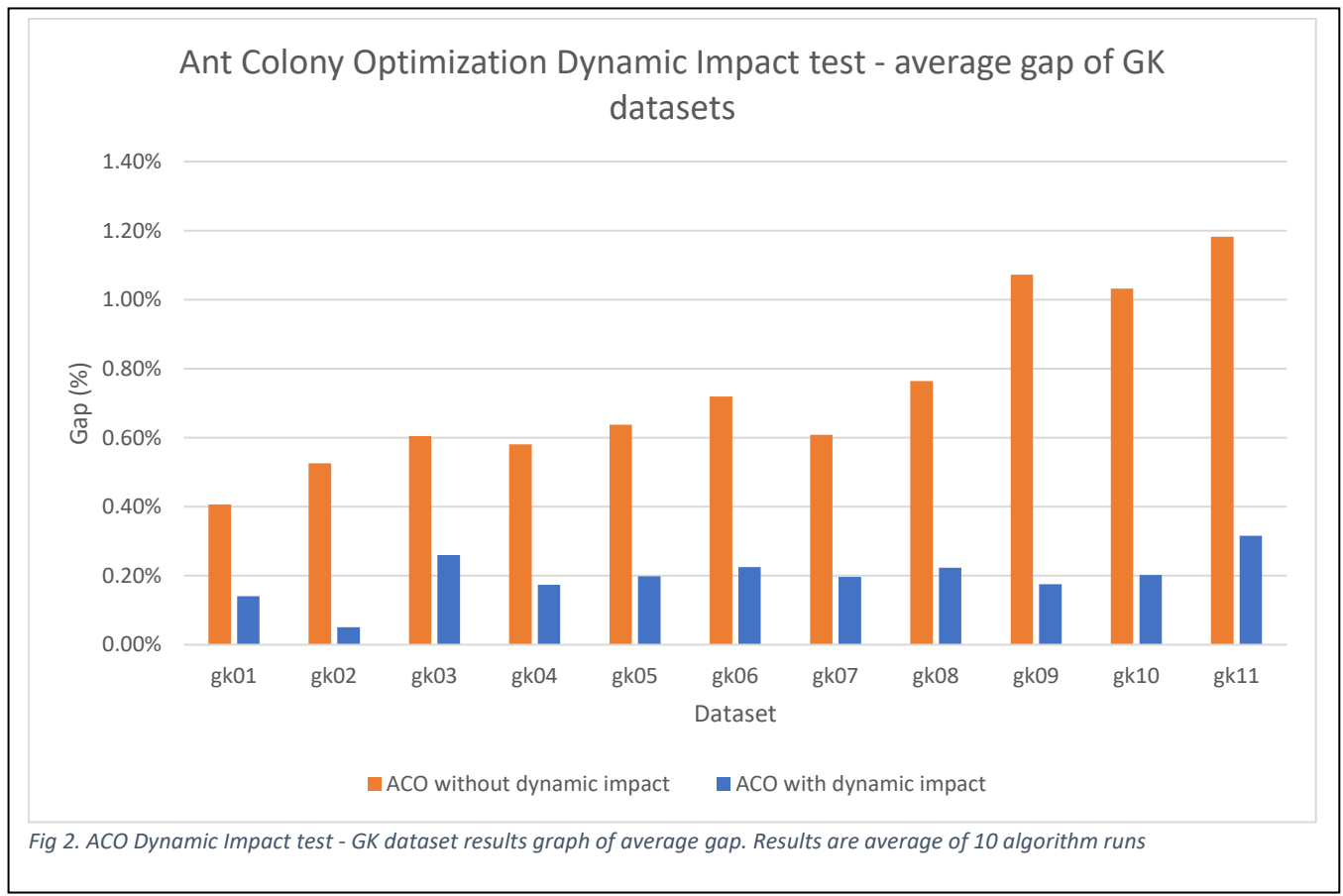

*Fig 2. ACO Dynamic Impact test - GK dataset results graph of average gap. Results are average of 10 algorithm runs*

Dynamic Impact has showed significant improvements solving Multi-dimensional knapsack problem. For small benchmark datasets Dynamic Impact solves all instances to optimal solution which is also a significant improvement in comparison to other published research. For large benchmark datasets Dynamic Impact can solve up to 10 times closer to known best or optimal value within same search efforts.

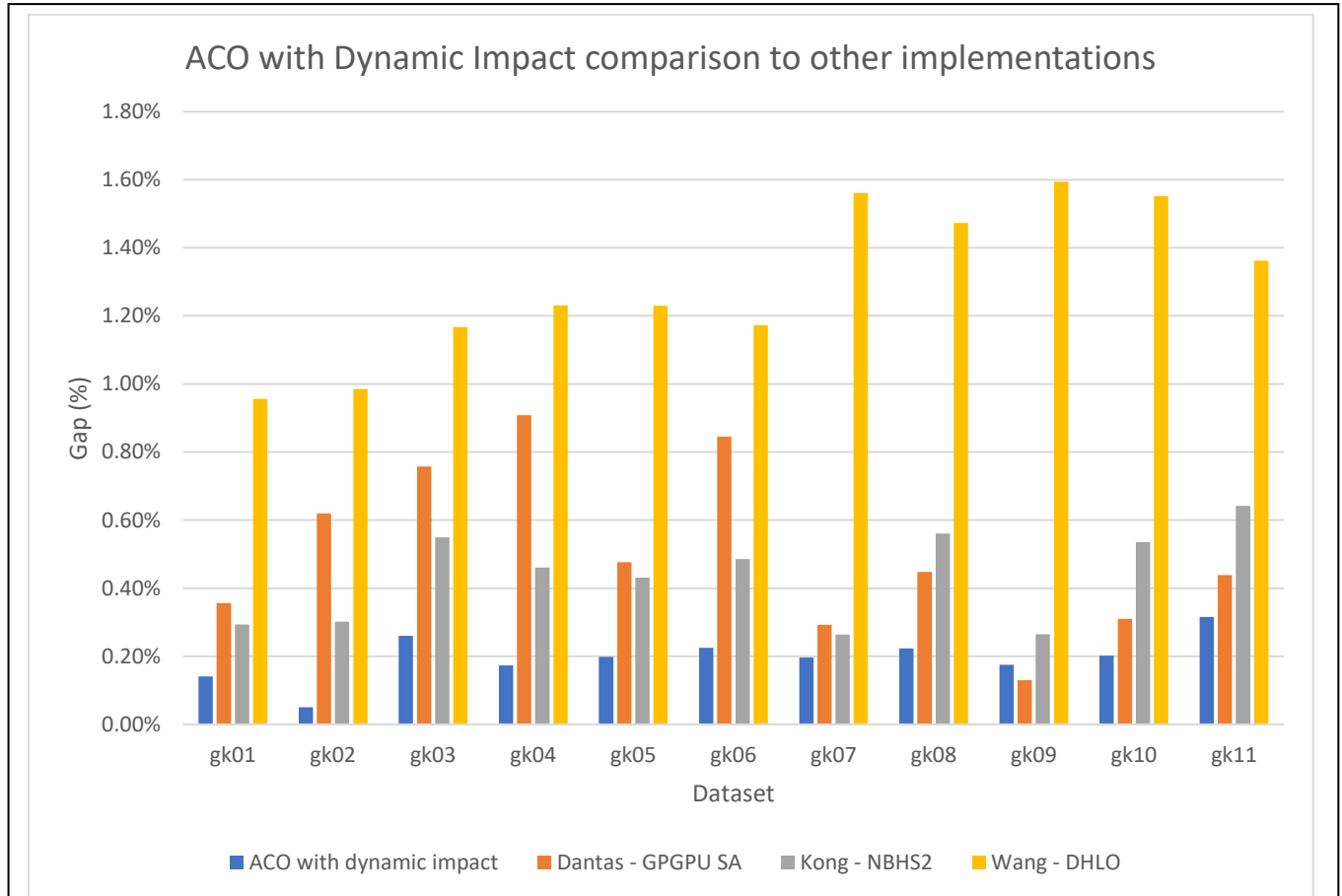

*Fig 3. ACO with Dynamic Impact comparison to other recently published GK dataset solution results.*

Table 6: MKP GK datasets. ACO results with Dynamic Impact are compared against ACO without Dynamic Impact as well as best performing other algorithm results taken from recently published papers.

| Dataset | problem size (N x M) | Best known profit | Average profit | | Average gap | | | | |
|---|---|---|---|---|---|---|---|---|---|
| | | | ACO without Dynamic Impact | ACO with Dynamic Impact | ACO without Dynamic Impact | ACO with Dynamic Impact | Dantas-GPGPU SA [44] 2018 | Kong-NBHS2 [45] 2015 | Wang-DHLO [46] 2017 |
| gk01 | 100 x 15 | 3766 | 3750.7 | 3760.7 | 0.41% | **0.14%** | 0.36% | 0.29% | 0.96% |
| gk02 | 100 x 25 | 3958 | 3937.2 | 3956 | 0.53% | **0.05%** | 0.62% | 0.30% | 0.99% |
| gk03 | 150 x 25 | 5656 | 5621.8 | 5641.3 | 0.60% | **0.26%** | 0.76% | 0.55% | 1.17% |
| gk04 | 150 x 50 | 5767 | 5733.5 | 5757 | 0.58% | **0.17%** | 0.91% | 0.46% | 1.23% |
| gk05 | 200 x 25 | 7560 | 7511.8 | 7545 | 0.64% | **0.20%** | 0.48% | 0.43% | 1.23% |
| gk06 | 200 x 50 | 7677 | 7621.8 | 7659.7 | 0.72% | **0.23%** | 0.85% | 0.49% | 1.17% |
| gk07 | 500 x 25 | 19221 | 19104.1 | 19183.29 | 0.61% | **0.20%** | 0.29% | 0.26% | 1.56% |
| gk08 | 500 x 50 | 18806 | 18662.3 | 18764 | 0.76% | **0.22%** | 0.45% | 0.56% | 1.47% |
| gk09 | 1500 x 25 | 58089 | 57466.1 | 57987.2 | 1.07% | 0.18% | **0.13%** | 0.27% | 1.59% |
| gk10 | 1500 x 50 | 57295 | 56703.9 | 57179.2 | 1.03% | **0.20%** | 0.31% | 0.54% | 1.55% |
| gk11 | 2500 x 100 | 95238 | 94111.6 | 94937.6 | 1.18% | **0.32%** | 0.44% | 0.64% | 1.36% |